# Simulated patient systems powered by large language model-based AI agents offer potential for transforming medical education

Huizi Yu, MS[1,2], Jiayan Zhou, PhD[3], Lingyao Li, PhD[1], Shan Chen, MS[4,5], Jack Gallifant, MBBS[4,5], Anye Shi, PhD[6], Jie Sun, MD[7], Xiang Li, MS[8], Jingxian He, BM[9], Wenyue Hua, PhD[10], Mingyu Jin, BS[10], Guang Chen, MD[11,12], Yang Zhou, MD[13], Zhao Li, MD[14], Trisha Gupte, MS[3], Ming-Li Chen, MD[3], Zahra Azizi, MD[3,15], Qi Dou, PhD[1], Bryan P. Yan, MD[1], Yanqiu Xing, MD[9], Yongfeng Zhang, PhD[10], Themistocles L. Assimes, MD[3], Danielle S. Bitterman, MD[4,5], Xin Ma, PhD[8*], Lin Lu, MD[16*], Lizhou Fan, PhD[1,2,4,5*]

1. The Chinese University of Hong Kong, Sha Tin, N.T., Hong Kong SAR, China
2. University of Michigan, Ann Arbor, MI, USA
3. Stanford University, Stanford, CA, USA
4. Artificial Intelligence in Medicine Program, Mass General Brigham, Boston, MA, USA
5. Harvard Medical School, Boston, MA, USA
6. Cornell University, Ithaca, NY, USA
7. Peking University Third Hospital, Beijing, China
8. School of Control Sciences and Engineering, Shandong University, Ji'nan, Shandong, China
9. Qilu Hospital of Shandong University, Ji'nan, Shandong, China
10. Rutgers University, New Brunswick, NJ, USA
11. Broad Institute of MIT and Harvard, Cambridge, MA, USA
12. The University of Hong Kong, Hong Kong, China
13. Chinese Academy of Medical Sciences and Peking Union Medical College, Beijing, China
14. Chinese Academy of Medical Sciences, Beijing, China
15. University of Ottawa, Ottawa, ON, Canada
16. Peking University Sixth Hospital, Beijing, China

†Corresponding authors:

Dr. Lizhou Fan (leofan@cuhk.edu.hk),
Dr. Lin Lu (linlu@bjmu.edu.cn), and
Dr. Xin Ma (maxin@sdu.edu.cn)

# Abstract

***Background:*** Simulated patient systems are vital in medical education and research, providing safe, integrative training environments and supporting clinical decision-making. Progressive Artificial Intelligence (AI) technologies, such as Large Language Models (LLM), could advance simulated patient systems by replicating medical conditions and patient-doctor interactions with high fidelity and low cost. However, effectiveness and trustworthiness remain challenging.
***Methods:*** We developed AIPatient, a simulated patient system powered by LLM-based AI agents. The system incorporates the Retrieval Augmented Generation (RAG) framework, powered by six task-specific LLM-based AI agents for complex reasoning. For simulation reality, the system is also powered by the AIPatient KG (Knowledge Graph), built with de-identified real patient data from the Medical Information Mart for Intensive Care (MIMIC)-III database.
***Results:*** Here we show that the system's accuracy in Electronic Health Record (EHR)-based medical Question Answering (QA), readability, robustness, and stability. Specifically, the system achieves a QA accuracy of 94.15% when all six agents, surpassing benchmarks with partial or no agent integration. Its knowledgebase demonstrates high validity (F1 score=0.89). Readability scores show median Flesch Reading Ease at 68.77 and median Flesch Kincaid Grade at 6.4, indicating accessibility to all medical professionals. Robustness and stability are confirmed with non-significant variance (ANOVA F-value=0.6126, $p > 0.1$; F-value=0.782, $p > 0.1$). A user study with medical students shows that AIPatient delivers high fidelity, usability, and educational value, matching or exceeding human-simulated patients in history-taking.
***Conclusions:*** Large language model–based simulated patient systems provide accurate, readable, and reliable medical encounters and demonstrates potential to transform medical education.

**Plain Language Summary:** Simulated patient systems are used to train medical students in realistic yet safe clinical settings. However, simulated patients played by human actors can be costly and limited in scale. In this study, we developed AIPatient, an artificial intelligence system that can simulate diverse patient interactions based on existing real electronic health records. We evaluated its accuracy, readability, and reliability, and compared it to an actor simulating human patients. The system achieved high accuracy in answering medical questions, produced clear and consistent responses, and was rated as highly realistic by students. These results suggest that AI-based simulated patients could make medical education more accessible, scalable, and consistent across different training environments.

# Introduction

Simulated patient (SP) systems, also referred to as virtual patient simulation systems, have become an essential tool in modern medical education and research [1]. These systems are designed to enhance integrative learning and evaluation by incorporating basic science objectives, simulating the outcomes of clinical decisions, and including diverse cases to improve cultural competency [2–5]. The applications of simulated patients are broad, spanning learning, teaching, and assessment [6,7].

While simulated patients are designed to be a low-risk and high-fidelity tool, there are unresolved concerns about their effectiveness and the trustworthiness. Current usability evaluation of simulated patients is usually subjective[8]. As a result, variations of medical students' familiarity of simulated patient systems could cause inaccuracy in SP-based medical skill evaluation [9]. Moreover, traditional evaluation of simulated patient systems often focuses on the fact-based accuracy and utilizes inter-rater reliability-based evaluation methods that are less generalizable [10]. For instance, in psychiatric education, inherent issues with utilizing simulated patients are highlighted in connection to the essence of empathy and the distinctively interpersonal nature of the field [11], which are traditionally not the focus of simulated patient system evaluation.

Large Language Models (LLM) are powerful Artificial Intelligence (AI) systems [12,13], especially renowned for their advanced reasoning capabilities [14,15] and medical application potentials [16–20]. LLM-based simulated patients have demonstrated promises in consistently replicating a wider range of medical conditions, simulating patient interactions, and mitigating logistical challenges associated with human actors. Some researchers have developed frameworks and methodologies for general medical education settings [21–23], while others have tailored these technologies to specific use cases such as psychiatry [24]. While these implementations are promising, there are still various concerns in effectiveness and trustworthiness of the systems, including (1) the lack of a large, comprehensive, and diverse patient profile database, (2) the need for fact-checking mechanisms to minimize hallucinations and ensure consistency, (3) the insufficient flexibility in taking on different personalities, and (4) the absence of a thorough evaluation framework.

Here, we introduce **AIPatient**, a simulated patient system with LLM-powered AI agents. It incorporates an advanced agentic workflow, *Reasoning Retrieval-Augmented Generation* (Reasoning RAG), and a reliable source of knowledge input, the *AIPatient Knowledge Graph* (AIPatient KG). The system processes realistic data from the Medical Information Mart for Intensive Care (MIMIC)-III and delivers verified information in a natural language format tailored to the user's needs, particularly aligning responses with patient personalities. It also ensures continuity in the interaction by summarizing and updating the conversation history throughout the process. In addition, the Reasoning RAG agentic workflow improves the traditional RAG framework [25], inserting step-by-step reasoning in between to advance the system's performance. In this study, AIPatient achieves high accuracy (94.15%) in Electronic Health Record (EHR)–based question answering, strong knowledge validity (F1 = 0.89), and readable, consistent responses (median Flesch Grade = 6.4). The system demonstrates robustness to varied question phrasing, stability across 32 personality types, and comparable or superior fidelity, usability, and educational value to human-simulated patients. These results show that AIPatient provides accurate, trustworthy, and scalable patient simulations, supporting its potential to transform medical education and clinical training.

# Methods

## Data

We create the patient profiles based on 1,500 sampled patient records from the Medical Information Mart for Intensive Care (MIMIC)-III database [26], containing de-identified real patient EHRs (**Supplementary Table 1**). This data use is approved by the PhysioNet Clinical Databases for credentialed access.

Using stratified sampling based on major diagnostic strata, the sampled of patient records accurately represents the diversity of the major diagnostic categories present in MIMIC-III (**Supplementary Fig. 2**). Among the EHRs, we focus on the structured data subset from the patient information table, admission information table, and vitals table, as well as unstructured data in discharge summaries.

## AIPatient KG construction with Named Entity Recognition

When constructing the AIPatient KG (**Supplementary Fig. 1**), we extract patients' symptoms, including medical history, vitals, allergies, social history and family history from the discharge summary using an LLM-based Named Entity Recognition (NER) approach. AIPatient KG has 1,500 patient-admission records, with a total of 15,441 nodes and 26,882 edges (**Supplementary Table 2**).

**Fig. 1** presents an example of NER-based data transformation for medical entities in discharge summaries. The relationships among these entities are stored in a graph database, Neo4j (AuraDB version 5) [27]. This data structure enables efficient storage, retrieval, and querying of complex relationships within the AIPatient KG.

## Reasoning RAG agentic workflow

**Fig. 2** shows the Reasoning RAG agentic workflow. The key stages in the workflow include *retrieval* with the Retrieval Agent and the KG Query Generation Agent, *reasoning* with the Abstraction Agent and the Checker Agent, and *generation* with the Rewrite Agent and the Summarization Agent. There are three initial inputs, including the AIPatient KG, a natural language query from the user, and the Conversation History between the user and the AIPatient. The first is only provided to the agents in the Retrieval Stage, and the latter two are universal inputs to all agents across the three stages. For example, **Fig. 3** shows a medical investigation that begins with a question input to the AIPatient system, "How long have you experienced soreness in your chest?" This question initiates a sequence of specialized agents that supports the *Reasoning RAG* process. Details of the operational flow of the multi-agent system can be found in **Supplementary Methods**.

## Evaluation and data labeling

To evaluate performance of the AIPatient system, we focus on its effectiveness and trustworthiness. Synthesizing the metrics in previous research of simulated patient systems and adapting them to the AIPatient system, we focus on five dimensions (**Table 1**).

We created two evaluation datasets to assess the five performance aspects. First, to evaluate LLM's performance for generating the knowledge graph, i.e., Knowledgebase Validity in the Named Entity Recognition (NER) task, we created a **gold-standard labeled dataset** with 100 cases labeled by expert medical practitioners [28]. Each case was labeled by two physicians and we perform the Intercoder Reliability Check by calculating span-level F1 scores, which stands at 0.79.

For the rest of the performance aspects, including Question and Answer (QA) Accuracy, Readability, Robustness and Stability, we developed a **medical QA conversation dataset** consisting of 524 questions [28]. We sampled 56 patient records from MIMIC-III and two

Natural Language Processing (NLP) researchers reviewed their discharge summaries. Based on this review, we formulated the questions and answers to focus on medical entities within the records.

**LLM selection and knowledgebase validity (NER) evaluation**

Data use agreement of MIMIC-III strictly prohibits the use of Protected Health Information (PHI) data with online services such as GPT. To conform with data use requirements, we use the Azure OpenAI service for GPT-family[29–31], Deepseek LLM[32], and Amazon Bedrock for Claude-family LLM[33], both not sharing data for model training [34]. In addition, we include two high-capacity open-source models—Qwen3-32B[35] and LLaMA-3-70B[36]—which are run entirely locally using the Ollama framework[37]. This setup ensures that all data processing remains on secure, institution-managed hardware without transmitting any content to external servers.

To evaluate model performance for downstream tasks such as clinical named entity recognition and question answering, we use the gold-standard labeled dataset and benchmarked 11 models: five Claude models [33] (Claude-3 Haiku, Claude-3-Sonnet, Claude-3.5 Sonnet, Claude-4- Sonnet and Claude-4-Opus), three GPT-family models[29–31] (GPT-4 Turbo, GPT-4o, and GPT-3.5 Turbo), and three open-source models including DeepSeek-V3 671B[32] , Qwen3-32B[35] and LLaMa-3 70B[36].

Our selection balances model diversity, performance, and compliance with data governance policies. The Claude and GPT models are well-established in biomedical information extraction and are benchmarked for their relevance to clinical tasks [38]. Open-source models are included to assess whether publicly accessible architectures can approach or match the performance of commercial LLMs, offering a pathway toward reproducibility and cost-efficient deployment[39]. We deliberately exclude very large reasoning models such as DeepSeek-R1, which are optimized for multi-step logical inference rather than structured extraction, and are computationally intensive for our token-efficient NER use case[40]. This selection strategy ensures that all models are aligned with privacy regulations while enabling a comprehensive and fair evaluation of current LLM capabilities in the clinical domain.

To perform NER tasks, we design individual prompts for extracting each type of medical entity. We use entity category-specific F1 scores to evaluate the knowledgebase validity. Additional implementation details are available in **Supplementary Methods**.

**QA accuracy evaluation through ablation studies**

In the QA conversation accuracy evaluation, we conduct ablation studies for the KG Query Generation Agent in different scenarios (1) on its own, (2) in combination with the Retrieval Agent, the Abstraction Agent, or both, and (3) using zero-shot or few-shot strategy (**Supplementary Fig. 3**). We conduct an ablation study with eight setups. Two researchers independently evaluate retrieved results against correct answers, providing binary ratings of correct or incorrect. An intercoder reliability check using Cohen's Kappa (0.92) ensures consistency [41]. Discrepancies are resolved by a third annotator, an experienced medical AI researcher. Model performance is assessed using accuracy rate, defined as the proportion of correctly answered questions out of the total.

**Readability evaluation**

To evaluate readability, we focus on the output of the Rewrite Agent, and employ two metrics the Flesch Reading Ease [42] and the Flesch-Kincaid Grade Level [43]. Both metrics are based on the Average Sentence Length (ASL, the number of words per sentence) and the Average Syllables per Word (ASW) to assess the readability of AIPatient and determine the ease of understanding for users. Here,

- **Flesch Reading Ease** =206.835−(1.015×ASL)−(84.6×ASW), where Higher scores indicate easier readability, with a score of 90-100 suggesting text easily understood by an average 11-year-old, and scores of 30-50 indicating college-level material.
- **Flesch-Kincaid Grade Level** = (0.39×ASL)+(11.8×ASW)−15.59, which estimates the U.S. school grade level required to understand the text, where lower grade levels correspond to simpler text, while higher grade levels reflect more complex material.

**Robustness (system) evaluation**

Each question in the QA conversation set is paraphrased three times to create a diverse pool of queries that maintain the original intent but differ in structure and vocabulary. To answer each paraphrased question, we invoke the Reasoning RAG framework and extract the required information from the knowledge graph. Finally, the robustness of AIPatient is assessed by comparing the accuracy of the responses generated from the paraphrased questions against the responses from the original questions (two sample t-test and ANOVA).

**Stability (personality) evaluation**

This evaluation focuses on ensuring that personality variations do not distort or omit essential clinical data, thereby maintaining consistency in the information presented across different simulated personalities. Using the Big Five personality framework [44], we generate 32 distinct personality types by combining different levels of the five major traits (**Supplementary Fig. 4**). These personality profiles are then integrated into the Rewrite Agent of the AIPatient system. To quantitatively assess these goals, we calculate the personality-induced data loss proportion, and employ ANOVA to test for statistical differences in the quality of information retention across the different personality profiles.

**Clinical Verification**

To evaluate the fidelity, usability, and educational effectiveness of the AIPatient system in a medical training context, we conducted a paired crossover user study comparing AIPatient with H-SPs in medical history taking. The user study involved twenty medical students and eight trained non-medical volunteers. A total of twenty simulated patient cases were selected from the MIMIC-III database, curated to represent common inpatient conditions based on ICD-9 classification. Each case was scripted to support both AI and human simulation consistently. All interactions were conducted via a unified web-based text interface, which masked the identity of the simulated patient to ensure experimental blinding.

After completing their assigned cases, participants filled out a structured Likert-scale questionnaire covering three core domains—fidelity, usability, and educational effectiveness. In addition, an Objective Structured Clinical Examination (OSCE)-style checklist was embedded to track whether participants elicited essential elements of a comprehensive medical history (e.g., chief complaint, medication review, psychosocial context)[45]. To complement the quantitative assessment, all participants also participated in semi-structured interviews following the interactions. These interviews were designed to capture in-depth qualitative feedback on system realism, engagement, and perceived training value. The study protocol was approved by the institutional review board (IRB Protocol Number: KYLL-202505-005), and all participants provided informed consent. This multi-method evaluation framework was designed to ensure a rigorous and ecologically valid assessment of AIPatient in comparison to traditional human-based simulation.

### Out of Distribution evaluation

We conduct Out of Distribution (OOD) performance analysis to evaluate the AIPatient system's ability to effectively process previously unseen or atypical clinical data. This analysis is performed using the CORAL (expert-Curated medical Oncology Reports to Advance Language model inference) dataset [46], which is an expert-curated collection of 40 de-identified oncology progress notes focused on breast and pancreatic cancer cases. Applying the same evaluation framework, we tested the AIPatient's performance in QA Accuracy, Readability, Robustness and Stability.

### Statistical analysis

In Robustness and Stability evaluation, we use ANOVA to statistically determine if there are significant differences in the system's performance across various input paraphrases and personality-infused responses. Two proportion t-tests are conducted to compare the accuracy of the system's responses between the original questions and their paraphrased counterparts, and between responses generated with and without personality traits. For the user experiment, two-sample t-tests were conducted to evaluate differences between AIPatient and Human Simulated Patients (H-SPs) across multiple questionnaire items measuring fidelity, usability, and educational value. A two-sided $p \leq 0.05$ was considered statistically significant. Statistical analyses were carried out using the statistical Python package in scipy (Scipy.org), Python version 3.9.16 (Python Software Foundation).

### System implementation

The AIPatient system was implemented in Python using modular agent classes that interact with closed-source large language models through secure cloud APIs (Azure OpenAI and Amazon Bedrock). Clinical entities such as symptoms, medical history, and allergies were extracted from MIMIC-III database for named entity recognition and stored as structured triples in a Neo4j graph database. The system operates through a sequential agentic pipeline, where each agent performs a distinct task—such as information retrieval, reasoning, summarization, or rewriting—via dedicated prompts and structured API calls. These agents communicate through shared JSON states, enabling multi-turn interactions that preserve memory and coherence. Multi-turn conversations were managed with prompt-embedded memory and a stateful dialogue manager.

# Results

### Evaluation of knowledgebase validity

Of the 11 large language model, we observe that the GPT-4-Turbo model has high knowledgebase validity, represented by model performance in NER task, with the highest overall F1 of 0.89 (**Fig. 4** & **Supplementary Table 3**). The GPT-4o model achieves the second-best performance with an F1 score of 0.75, followed by the Claude Sonnet and Opus models (Sonnet 3.5, Sonnet 4, and Opus 4), which perform the third best with an average F1 score of 0.73. We note that GPT-4o and GPT-4-Turbo models tend to have higher precision across most categories, which is beneficial for reducing the noise in data processing and knowledge graph construction. Additionally, we observe the GPT-family models specifically excels in extracting Allergies, where the older version of Claude models suffer (Claude-3-haiku and Claude-3-sonnet). Open-source models, including DeepSeek, Qwen3-32B, and LLaMA3-70B, generally show lower overall F1, precision, and recall scores compared to their commercial counterparts. Based on these results, we use the GPT-4 Turbo model to construct the final version of AIPatient KG. Specific numerical results for F1, Precision and Recall are presented in **Supplementary Tables 3-5**.

### Analysis of AI agent abilities via QA accuracy ablation studies

In **Table 2**, we highlight ablation studies results. The setup with all agents and few-shot learning achieves the highest accuracy in most entity categories, with 94.15% overall accuracy. The baseline without the AIPatient KG and Reasoning RAG performs worse, particularly in Family and Social History, where accuracy drops to 13.33%. Results of additional entity categories (Admission, Patient, Allergy, Vitals) are presented in **Supplementary Table 6**, where all LLM achieves a high accuracy rate regardless of the setup chosen.

We further compare different models under the best-performing setup (all agents with few-shot learning) in **Table 3**. GPT-4-Turbo achieves the highest overall accuracy (94.15%), followed closely by Claude-4-Opus (90.80%) and GPT-4o (89.02%), with strong performance across all entity categories. Open-source models such as Deepseek-v3-671b and Qwen-3-32b show promising results in Medical History (79.31% and 77.27%, respectively), but consistently underperform in most entity categories. These findings underscore the superior reliability of advanced proprietary models as the backbone in handling complex question answering tasks.

### Investigation of system's readability, robustness, and stability

We use algorithmic methods and existing metrics to investigate system readability, system robustness, and stability charted by personality variants (see **Methods** for details).

For Readability, the AIPatient system presents Flesch Reading Ease scores ranging from 10.91 to 99.23 (median 68.77, **Fig. 5(a)**) and Flesch-Kincaid Grade Level peaking at the sixth-grade level (median grade level 6.4, **Fig. 5(b)**). The median Flesch Reading Ease score is within the aim score between 60 and 70; Flesch-Kincaid Grade Level is below 8, the required grade level for the general public readership [42,43]. These values reflect that the system can simulate naturalistic patient responses.

ANOVA and two-sample t-tests (**Supplementary Tables 7 & 8**) show that there is no significant effect of QA conversation paraphrasing on overall response accuracy ($F = 0.6126$, $p = 0.5420$), indicating robustness to question phrasing changes. However, in the Medical History category, accuracy is significantly affected ($F = 5.3038$, $p = 0.00589$), likely due to the complexity of such queries. No significant differences are found in the Symptom and Family and Social History categories, demonstrating resilience to input phrasing variations in these areas.

Across 32 personality groups, the median data loss is 2% (range: 0%–5.88%), indicating consistently low variability. For the Overall dataset, personality groups show no significant impact on performance (**Supplementary Table 9**). Similarly, no significant differences are observed for Symptom ($F = 1.1104$, $p = 0.3230$) and Family and Social History ($F = 0.6774$, $p = 0.9024$) categories. These results confirm the system's stability and consistency across diverse simulated personalities.

### Evaluation of the AIPatient system to human-simulated patients in a clinical education setting

We design a paired crossover experiment to evaluate and compare the fidelity, usability, and educational effectiveness of AIPatient and human-simulated patients (H-SPs) in a medical training context. The study recruits 20 medical students and 10 trained non-medical volunteers (H-SPs). Each medical student completes two patient cases, engaging in a total of four interactions—one with AIPatient and one with a H-SP for each case. To preserve experimental blinding and minimize bias, all interactions are conducted through a unified online text-based interface that masks the identity of the simulated patient. Following all interactions, participants complete a structured Likert-scale questionnaire, an OSCE-style checklist, and participates in a semi-structured interview to capture qualitative insights on their experience with AIPatient and H-SPs. Details of questionnaire and study design are available in **Supplementary Table 13** and **Supplementary Methods** respectively.

Across most dimensions, AIPatient performs comparably or better than H-SPs (**Fig. 6(a)** and **Table 4**). We conduct two-sample t-tests (two-sided) to compare the mean scores across fidelity, usability, and educational effectiveness domains. In the domain of fidelity, AIPatient scores higher in both case script adherence (mean = 4.32 vs. 4.08) and alignment with the intended medical condition (4.24 vs. 3.90, $t = 1.77$, *$p < 0.1$), with a particularly notable advantage in emotional realism (4.37 vs. 3.74, $t = 3.41$, **$p < 0.01$). AIPatient is also perceived as more coherent (4.32 vs. 4.08) and clinically relevant (4.27 vs. 4.23). Usability metrics also favors AIPatient, which scores higher in ease of interaction (4.20 vs. 3.79), technical reliability (4.39 vs. 3.79, $t = 2.68$, ***$p < 0.01$), and integration potential (4.02 vs. 3.92). In the domain of educational effectiveness, AIPatient shows advantages in supporting diagnostic accuracy (4.27 vs. 3.87) and improving clinical reasoning skills (4.41 vs. 3.97, $t = 2.19$, **$p < 0.05$). Additionally, the OSCE-style checklist, designed to assess students' ability to gather essential clinical information, shows that AIPatient matches or exceeds human-simulated patients in most areas—demonstrating strong performance in supporting clinical reasoning and information elicitation (**Fig. 6(b)**).

Interview themes echos the above quantitative findings. Medical students find AIPatient interactions to be emotionally expressive, pedagogically valuable, and efficient. Its consistent personality traits and rapid responses enable streamlined case engagement. However, some note overly verbose outputs and limited responsiveness to non-standard queries. Overall, AIPatient is perceived as a usable and high-fidelity training tool with promising applicability to medical education. In-depth quantitative and qualitative insights are provided in **Supplementary Discussions**.

### Evaluation of AIPatient on out of distribution (OOD) data

The AIPatient system using CORAL data achieves comparable performance to the MIMIC-III version. The highest QA accuracy (81.04%) occurs with all agents and few-shot learning (**Supplementary Table 10**). AIPatient's QA accuracy is lower with CORAL data, likely due to the specialized language and complexity of oncology reports. For readability, the system achieves a median Flesch Reading Ease of 70.6 and a Flesch-Kincaid Grade Level of 6.8, consistent with the MIMIC-III-powered system (**Supplementary Fig. 5**). ANOVA tests on robustness and stability show no significant differences in paraphrasing and different personality groups respectively (**Supplementary Tables 11 & 12**). These results highlight the AIPatient system's adaptability and robust performance across datasets and testing conditions, handling complex medical narratives effectively.

# Discussion

In this research, we explore how the AIPatient system, leveraging LLM-based AI agents, enhances the performance of simulated patient systems. Our quantitative results underscore the advantages of integrating interactive agents in LLM-based reasoning workflow to effectively and reliably support medical investigations. The ablation study on AI agents demonstrates that multi-agent design can enhance overall system performance, outperforming individual LLM models in medical question answering. The subsequent metric-based analysis further indicates that the AIPatient system presents balanced effectiveness and reliability regarding readability, robustness, and stability. Evidence from this study suggests a potential avenue for incorporating LLM-based AI agents in intelligent medical systems.

A key feature of the AIPatient system is its workflow powered by LLM-based agents. The sequence, interaction, and collaboration of the agents together enhance the system's realism and accuracy in medical question answering. Unlike traditional one-off prompt responses, the Reasoning RAG framework employs self-reflection, abstraction, and iterative checking to deliver reliable answers. These features enable agents to collaborate dynamically, distilling complex medical data and refining responses through iterative processes. This framework also directly addresses risk of bias by reducing information omission or over-reliance on single-step outputs. The agent collaboration ensures that clinical queries are checked, abstracted, and cross-verified before generating a final answer, thereby improving completeness and consistency in medical reasoning.

Regarding effectiveness, the system achieves an accuracy of 94.15% by combining all six agents and using intelligent prompt design. Few-shot learning strategies add depth to the system, improving QA accuracy by 11.1% on average**.** The Readability aspect of the AIPatient system further confirms that the correct and accurate generated patient response is easy to understand – the median value for Flesch Reading Ease and Flesch-Kincaid Grade Level mimic the readability of the fiction novel *Harry Potter*. This level of reading complexity is suitable for most medical investigators, such as medical students.

Trustworthiness is another vital aspect that verifies AIPatient as a responsible AI system. First, the system exhibits strong robustness to the inherent variability in LLM outputs, as paraphrasing medical questions does not significantly affect overall QA accuracy, indicating that the system can consistently retrieve accurate information despite variations in question phrasing. Second, the system is stable, capable of integrating distinct personality traits into simulated patient interactions without compromising the accuracy or integrity of medical information. These findings highlight the need for medical AI systems to accommodate linguistic variability and patient diversity, ensuring reliable performance across different phrasings, personality traits, and clinical contexts.

Moreover, our approach represents an advancement in LLM-based patient simulation by combining structured Electronic Health Records (EHR) processing with multi-agent systems to support human-like interactions. The construction of the AIPatient KG prototypes a large-scale NER-based EHR processing. Not only does this approach allow for clear data organization, but it also enables sophisticated queries and flexible schema updates, making the system adaptable to evolving medical knowledge and patient needs.

To further assess AIPatient's real-world applicability, we benchmark its computational efficiency and cost across popular LLMs (**Supplementary Discussion**). Proprietary models such as GPT-4o and Claude-3-Haiku show strong performance in both speed and affordability, making them viable for real-time clinical simulation. GPT-4-Turbo emerges as the most balanced model, supporting its use as the system's backbone. This analysis underscores the system's scalability and offers guidance for selecting models suited to practical deployment needs. Additionally, the modular agentic design of AIPatient allows new models to be integrated at the component level without full system re-evaluation, supporting flexible upgrades and targeted benchmarking as better models emerge.

Our user evaluation with medical students further highlights AIPatient's educational value. In a paired crossover study with H-SPs, AIPatient matches or outperforms H-SPs across most metrics—including emotional realism, reliability, and learner satisfaction—demonstrating its ability to deliver high-quality, pedagogically effective simulations. Qualitative feedback echoes these strengths, with students commenting on AIPatient's expressiveness, speed, and consistency. Some areas for improvement are identified, such as managing verbosity and increasing flexibility in handling non-standard queries, pointing to future opportunities for system refinement.

Our study fits into the broader literature by addressing a critical gap in using AI for patient simulation. While previous work has explored simulation systems for hospital management or behavioral modeling [21–23], the AIPatient system uniquely applies multi-agent workflows with an emphasis on reasoning to simulate patient interactions. The AIPatient system is also one of the first that incorporates real, de-identified patient information and human-algorithm collaborative system evaluation, outperforming the precedent patient simulation systems that are mostly based on a small number of fixed cases and are manually evaluated in domain-specific manners [24]. Our system also demonstrates how AI agents can collaborate as a system to tackle complex medical reasoning tasks, as one of the first implementations of LLM-based agentic systems in medicine and healthcare [47], providing an innovative tool for both education and healthcare delivery.

Despite its achievements, this present study has several limitations. First, the reliance on discharge notes from MIMIC-III restricts the diversity of patient cases, and the homogeneous population it represents limits generalizability. Expanding the database to include outpatient, inpatient, and long-term care settings, as well as populations from varied demographics, will address these issues. Second, the system shows varying performance across different entity categories, highlighting areas like social history as needing further refinement. In addition, while the current version of AIPatient incorporates diverse personality traits to simulate variations in patient behavior, it does not yet explicitly model broader social determinants of health (SDoH) such as socioeconomic status, education level, or living conditions. These factors can significantly influence how patients communicate symptoms, adhere to care plans, and engage with clinicians.

Future research should explore the ethical, psychological, and professional dimensions of implementing generative AI, especially LLM, in clinical settings. Feedback from medical trainees and professionals in the user study highlights the system's strengths—such as emotional realism, consistency, and perceived educational value—while also identifying areas for refinement, including managing verbosity and improving flexibility with non-standard queries. These insights demonstrate the importance of stakeholder involvement in guiding system development. Future research should expand this participatory approach by engaging a broader range of users, including clinicians and patients from diverse backgrounds, to further align AIPatient with real-world expectations and values. Additionally, further iterations of AIPatient can integrate structured SDoH data to support more inclusive and equity-aware training scenarios. This will further enhance the system's ability to prepare trainees for real-world variability in patient presentations and healthcare access. Overall, the AIPatient system exemplifies the promise of responsible AI in bridging the gap between clinical training and patient care, laying a foundation for more inclusive, accurate, and effective medical simulation systems.

The findings from our study demonstrate a simulated patient system powered by large language model-based AI agents. This offers potential for transforming medical education by providing trainees with a realistic, scalable, and accessible tool to practice clinical reasoning. Beyond education, the system has implications for healthcare practice, where it can support clinicians by simulating rare or complex cases, helping to refine diagnostic approaches or evaluate treatment options. Its adaptability to new data types, such as imaging and multimodal inputs, further extends its utility, promising a more holistic simulation of patient care.

# Ethics

Our research was based on the MIMIC-III dataset [26] and the CORAL dataset [46], accessed through PhysioNet [48]. Prior training of CITI Data or Specimens Only Research (Record Number: 59460661) was completed on November 2, 2023. LLM were used in compliance with PhysioNet standard, including using the GPT models on Azure and the Claude models on Amazon Bedrock for data privacy.

The user study involving medical students and volunteers was approved by the Institutional Review Board of Qilu Hospital of Shandong University (IRB Protocol Number: KYLL-202505-005). All participants provided informed consent prior to participation. All procedures involving human participants were conducted in accordance with the Declaration of Helsinki.

# Data Availability

Our data, AIPatient KG and AIPatient KG-CORAL, as well as the corresponding medical QA datasets, are available on PhysioNet [28]. These de-identified data may be made available upon reasonable request via a proposal-based application process through PhysioNet. Data usage agreement and training facilitated by PhysioNet are required for the application. Source data including bootstrap outputs used to generate **Fig. 4,** readability scores for **Fig. 5,** and medical-student scoring data for **Fig. 6** are available on Figshare[49]. We cannot release the full clinical-text–derived datasets openly. Upstream licenses for MIMIC/PhysioNet credentialed data prohibit onward redistribution and the use of third-party online services; derivatives that could enable re-identification (e.g., note-aligned triples or QA pairs) must remain under controlled access. Although de-identified, narrative notes still carry non-zero re-identification risk, which is why access is managed rather than public. Portions of our work also use CORAL (de-identified oncology notes). Its availability and reuse are governed by the original owners/publisher; consequently, our CORAL-based graph and QA derivatives are released only within the same controlled-access framework, and we cannot repost CORAL itself.

URL link for AIPatient KG and AIPatient KG-CORAL, and corresponding medical QA datasets: https://physionet.org/content/aipatient-kg/1.0.0/
URL link for Fig. 4, Fig. 5, Fig. 6 raw data: https://doi.org/10.6084/m9.figshare.30327508.v2
URL link for MIMIC-III raw data: https://physionet.org/content/mimiciii/1.4/
URL link for CORAL raw data: https://physionet.org/content/curated-oncology-reports/1.0/

# Code Availability

The code for data processing, system construction, experiments, and system demonstration is publicly available via Zenodo[50]: http://dx.doi.org/10.5281/ZENODO.14583946. The most up-to-date version is available on GitHub: https://github.com/huiziy/AIPatient.

# Figure Legend

**Fig. 1**: Data transformation of EHRs from **(a)** raw discharge notes (with extracted entities) to **(b)** constructed knowledge graph (through NER). Here, legend indicates medical entity and color mapping and only a subset of symptoms and medical history is presented in the figure for clarity. The patient's family medical history is described as "Both parents died from CVA's." NER extracts "both parents" as Family Member, and "cerebrovascular accident (CVA)" as Medical History. A relationship of "HAS_FAMILY_MEMBER" is established between the Patient node and Family Member node (both parents), and "HAS_MEDICAL_HISTORY" is established between the Family Member node and the Medical History node (CVA).

**Fig. 2**. Reasoning RAG agentic workflow is the AIPatient system's processing backbone, comprising three key stages: retrieval, reasoning, and generation. It first retrieves relevant information from the knowledge graph (Retrieval Agent and KG Query Generation Agent), then applies contextual reasoning to reduce hallucinations (Abstraction Agent and Checker Agent), and finally generates natural language responses based on conversation continuity and tailored to the perceived patient personality (Rewrite Agent and Summarization Agent). Created by the authors using built-in Microsoft PowerPoint icons and shapes.

**Fig. 3** An AIPatient input and output example with the Reasoning RAG inner process. The user asks about chest soreness duration. The Abstraction Agent reformulates the intent ("duration of symptom"), the KG Query Generation Agent composes a Cypher query to the patient-specific knowledge graph, and the Retrieval Agent returns linked nodes (Symptom: soreness in chest; Duration: 6–8 months). The Checker Agent verifies that the result answers the abstracted question or triggers a paraphrase-and-retry loop. Conditioned on a personality profile, the Rewrite Agent generates an empathetic first-person reply, and the Summarization Agent updates the running context. Together, these agents maintain conversation continuity and reduce hallucinations while producing the final natural-language response. Created by the authors using built-in Microsoft PowerPoint icons and shapes.

**Fig. 4.** Boxplot comparison of F1 Score, Precision, and Recall across different entity categories for 11 large language models. Each distribution is constructed from 1,000 bootstrap resamples per metric and model. Boxes show the 25th–75th percentiles with the median line; whiskers extend to the most extreme data points within 1.5×IQR from the first and third quartiles. (a) F1 Score comparison across entity categories, with boxes representing the F1 score and color indicating the large language models. GPT-4-Turbo generally leads in F1 scores, suggesting a robust balance of precision and recall. (b) Precision Score comparison across entity categories, with boxes representing the precision and color indicating the large language models. GPT-4o and GPT-4-Turbo excel in precision across most categories, making them suitable for applications where precision is prioritized to minimize false positives. (c) Recall Score comparison across entity categories, with boxes representing the recall and color indicating the large language models. Comparison shows high recall rates of GPT-3.5-Turbo, especially in the Medical History category.

**Fig. 5**: Distribution of the Flesch Reading Ease and the Flesch-Kincaid Grade Level of AIPatient system outputs. Red dashed vertical lines indicate the median in each panel (Flesch Reading Ease = 68.77; Flesch-Kincaid Grade Level = 6.4), showing that most responses cluster around middle-school readability.

**Fig. 6**. **Evaluation of Simulated Patient System Using OSCE Checklist and User Experience Questionnaire.**
**Panel (a)** Medical students rated their experience with the AIPatient and H-SPs using a 5-point Likert scale across multiple dimensions including coherence, clinical relevance, and emotional realism.
**Panel (b)** Comparison of information collected in the Objective Structured Clinical Examination (OSCE) checklist between AIPatient and H-SPs, showing the proportion of key clinical information items successfully elicited during simulated interactions.

# Tables

**Table 1**. Evaluation framework

| Performance aspect | Evaluation dimension | Evaluation by | Metrics |
|---|---|---|---|
| Effectiveness | Knowledgebase validity (NER) | Medical doctors | F1 |
| | QA accuracy (conversation) | Researchers | Accuracy |
| | Readability | Algorithm | Flesch Reading Ease, Flesch-Kincaid Grade Level |
| Trustworthiness | Robustness (system) | Researchers | Accuracy, ANOVA |
| | Stability (personality) | Researchers | Accuracy, ANOVA |

**Table 2** Ablation Studies Result[1] by Entity Category (Differentiating Set[2])

| Few Shot | Retrieval Agent | Abstraction Agent | Overall | Symptom Group | Medical History | Family and Social History |
|---|---|---|---|---|---|---|
| ✓ | ✓ | ✓ | **94.15%**[3] | **91.20%** | **87.10%** | **85.56%** |
| ✓ | ✓ | | 92.60% | 89.68% | 83.87% | 78.89% |
| ✓ | | ✓ | 93.80% | 90.48% | 83.87% | **85.56%** |
| ✓ | | | 92.94% | 90.48% | 69.35% | 82.22% |
| | ✓ | ✓ | 81.41% | 85.71% | 25.81% | 60.00% |
| | ✓ | | 81.93% | 84.92% | 27.42% | 58.89% |
| | | ✓ | 83.13% | 87.20% | 30.65% | 64.44% |
| Only with *KG Query Generation Agent* | | | 82.62% | 88.80% | 25.81% | 60.00% |
| Without *Reasoning RAG* & Without *AIPatient KG* | | | 68.94% | 64.29% | 53.45% | 13.33% |

[1]All results are with *AIPatientKG* as the input, unless specified otherwise.
[2]This differentiating set excludes admission, patients, allergies and vitals, for which we observe 100% or close to 100% accuracy regardless of setup.
[3]Highest accuracy in each category is in bold.

**Table 3** QA Accuracy Result[1] by Model[2]

| Model Type | Model Name | Overall | Symptom Group | Medical History | Family and Social History |
|---|---|---|---|---|---|
| Closed-Source | Claude-3-Haiku | 72.25% | 68.00% | 40.91% | 56.76% |
| | Claude-3.5-Sonnet | 83.82% | 88.00% | 86.36% | 48.65% |
| | Claude-4-Sonnet | 86.13% | 68.00% | 86.36% | 64.86% |
| | Claude-4-Opus | 90.80% | 88.00% | 86.36% | 78.38% |
| | GPT-3.5-Turbo[3] | 55.49% | 60.00% | 36.36% | 60.00% |
| | GPT-4o | 89.02% | 92.00% | 81.82% | 64.86% |
| | GPT-4-Turbo | 94.15%[3] | 91.20% | 87.10% | 85.56% |
| Open-Source | Deepseek-v3-671b | 80.81% | 63.64% | 79.31% | 62.50% |
| | Llama-3-70b | 75.72% | 84.00% | 36.36% | 43.24% |
| | Qwen-3-32b | 78.61% | 84.00% | 77.27% | 40.54% |

[1]All model tested using setup with all agents and few-shot learning.
[2]Claude-3-Sonnet model is removed due to model deprecation.
[3]We observe that GPT-3.5-Turbo performs poorly in the Cypher query generation task, suggesting limitations in structured reasoning and translating natural language into formal queries.

**Table 4 Two-sample t-test results comparing AIPatient and H-SP**

| *Metric* | *Question* | *t-statistics* |
|---|---|---|
| ***Fidelity*** | | |
| *Role/Text Adherence* | The SP followed the case script without contradictions. | 0.57 |
| | The SP’s responses matched the intended medical condition. | 1.77* |
| *Contextual Appropriateness* | The SP’s responses felt natural and relevant to my questions. | 1.10 |
| *Emotional Realism* | The SP displayed believable emotions (e.g., pain, anxiety). | 3.02 *** |
| *Coherence/Consistency* | The SP’s dialogue was coherent (no abrupt shifts). | 1.23 |
| *Response Quality* | The SP’s answers were directly relevant to clinical questions. | 0.17 |
| ***Usability*** | | |
| *Ease of Use* | Interacting with this SP required minimal effort. | 1.62 |
| | I encountered no technical difficulties (e.g., delays). | 2.68 *** |
| *Feasibility/Scalability* | This SP could be easily integrated into our training program. | 0.47 |
| ***Effectiveness*** | | |
| *Diagnostic Accuracy* | I have reached a preliminary diagnosis at the end. | 1.59 |
| *Learner Satisfaction* | This session improved my clinical reasoning skills. | 2.19 ** |

** : $p < 0.1$, ** : $p < 0.05$, *** : $p < 0.01$; reported p values are from two-sided tests.*

# Acknowledgments

The authors acknowledge the following funding sources: the Chinese University of Hong Kong Vice-Chancellor Early Career Professorship Scheme (L.F.), National Natural Science Foundation of China (Grant No. 62573271) (X.M.), and Major Basic Research Project of Shandong Provincial Natural Science Foundation (Grant No. ZR2025ZD28) (X.M. and L.F.), and NIH-USA R01CA294033 (S.C., J.G., D.S.B., and L.F.). The content is solely the responsibility of the authors and does not necessarily represent the official views of universities or hospital authorities. The authors acknowledge the help from Dr. Libby Hemphill for the initial data acquisition.

# Author contributions

H.Y.: conceptualization, data curation, formal analysis, investigation, methodology, visualization, writing—original draft, writing—review & editing. J.Z. and L.Y.L: data curation, formal analysis, investigation, methodology. S.C., J.G., A.S., J.S., X.L., W.H., M.J.: data curation, investigation, writing—review & editing. G.C., Y.Zhou, Z.L., T.G., M-L.C., and Z.A.: data curation, investigation, methodology. Y.Zhang, T.L.A., and D.S.B.: conceptualization, methodology, writing—review & editing. X.M. and L.L.: conceptualization, investigation, supervision, writing—review & editing. L.F.: funding acquisition, conceptualization, data curation, formal analysis, investigation, methodology, supervision, writing—original draft, writing—review & editing.

# Competing interests

D.S.B.: Associate Editor of Radiation Oncology, HemOnc.org (no financial compensation, unrelated to this work); funding from American Association for Cancer Research (unrelated to this work). All other authors declare they have no competing interests.